# Application-Driven AI Paradigm for Human Action Recognition

Zezhou Chen
ChinaUnicom
chenzz51@chinaunicom.cn

Yajie Cui
ChinaUnicom
cuiyj62@chinaunicom.cn

Kaikai Zhao
ChinaUnicom
zhaokk3@chinaunicom.cn

Zhaoxiang Liu*
ChinaUnicom
liuzx178@chinaunicom.cn

Shiguo Lian*
ChinaUnicom
liansg@chinaunicom.cn

**Abstract**

*Human action recognition in computer vision has been widely studied in recent years. However, most algorithms consider only certain action specially with even high computational cost. That is not suitable for practical applications with multiple actions to be identified with low computational cost. To meet various application scenarios, this paper presents a unified human action recognition framework composed of two modules, i.e., multi-form human detection and corresponding action classification. Among them, an open-source dataset is constructed to train a multi-form human detection model that distinguishes a human being's whole body, upper body or part body, and the followed action classification model is adopted to recognize such action as falling, sleeping or on-duty, etc. Some experimental results show that the unified framework is effective for various application scenarios. It is expected to be a new application-driven AI paradigm for human action recognition.*

## 1. Introduction

In actual applications, various scenarios require the monitoring of human actions. In scenarios involving safety supervision, surveillance cameras obtain image information of the safety supervisors' work area to monitor whether safety supervisors are on duty or even whether they are sleeping. In some scenarios, the number of personnel needs to be limited, such as statistics on workers in the work areas of the factory assembly line where the number of personnel should within a compliant range. In the environment of chemical plants, fall detection can find the unnormal human body in time to alert possible gas leakage accidents. Therefore, human action recognition has important research significance for reducing safety accidents in many scenarios.

There are many studies on human action recognition. Two-stream RNN/LSTM framework [44, 47, 49] takes different input features extracted from the RGB videos for Human Action Recognition (HAR) and gets human action results through fusion strategies. However, the computational complexity of the two-stream framework is huge compared to a single CNN framework, and it relies on continuous image input. 3D CNN framework [39, 40, 54] extends 2D CNNs to 3D structures, to simultaneously model the spatial and temporal context information in videos. But this kind of algorithm is also computationally intensive and relies also on continuous image input. Skeleton based methods [27, 30, 34, 63] extract skeleton sequences to encode the trajectories of human body joints, which characterize informative human motions. However, these algorithms are not only computationally intensive, but also unstable in actual monitoring scenarios. Numerous works [9, 21, 26] has made great contributions to the improvement of dl-based object detection algorithms. Regression based algorithm [17] directly estimates the count of pedestrian. The method [41] detects fall actions via SVM. And skeleton based and human key-points based algorithms [37, 58] are proposed for fall detection.

Generally, human action recognition depends on captured visual information sincerely. As shown in Figure 1, some scenes need the whole body to make a decision, e.g., fall detection, and some scenes need the upper body to make a decision, e.g., sleeping detection, while some other scenes need only part of body to make a decision, e.g., person counting for on-duty detection. Now, most algorithms consider each kind of scene independently, while a unified framework is expected to support various scenes simultaneously. Additionally, inspired by the fact that humans may need a single instance of visual information to recognize human actions, it is possible to realize accurate identification with a single frame of a video, which may reduce the computational cost on temporal sequence processing.

---

*corresponding author

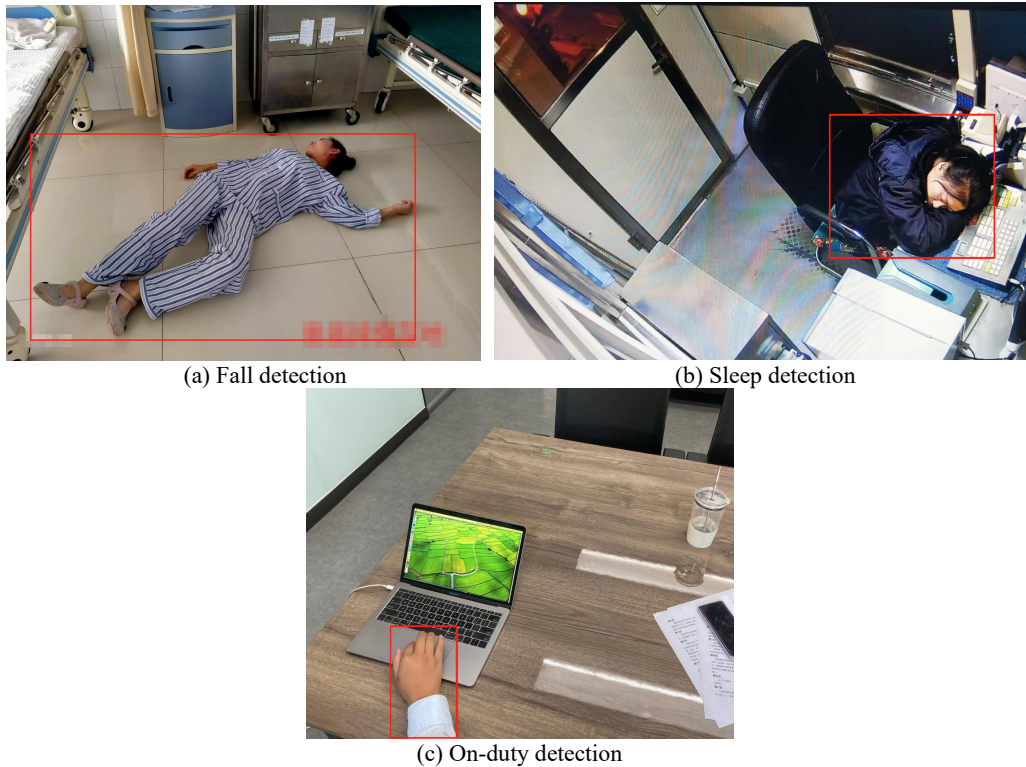

(a) Fall detection  (b) Sleep detection
(c) On-duty detection

Figure 1. Human action recognition in different scenarios depends on different visual information including the whole body, upper body or part of body.

To address the above mentioned issues, we propose a unified human action recognition framework composed of multi-form human detection module and action classification module. With a single image as input, the multi-form human detection module localizes the whole body, upper body and parts of body simultaneously with various labels. And then, the localized human bodies are fed to corresponding action classifications which identify corresponding actions in the image. The scheme accepts input images frame by frame, and assigns action labels based on individual frame. Combining the results for a time period, the human actions are predicted. The contributions of paper include mainly two aspects: 1) A unified human action recognition framework is proposed as a new application-driven AI paradigm; 2) The multi-form human detection model is proposed and the corresponding dataset is open-sourced; 3) Some experiments are done to show the proposed paradigm's effectiveness.

The rest of the paper is arranged as follows. In Section 2, some related works are introduced. The proposed AI paradigm is presented in detail in Section 3. In Section 4, the constructed multi-form human detection dataset is presented in detail. Some experiments are done and results are given to show the effectiveness of the proposed AI paradigm in Section 5. Finally, in Section 6, some conclusions are drawn.

## 2. Related Works

Human Action Recognition (HAR), i.e., recognizing and understanding human actions, is crucial for a number of real-world applications. It can be used in visual surveillance systems [20] to identify dangerous human activities, and autonomous navigation systems [25] to perceive human behaviors for safe operation. Besides, it is important for a number of other applications, e.g., video retrieval.

In the early days, most of the works focused on using RGB or gray-scale videos as input for HAR [42], due to their popularity and easy access. Recent years have witnessed an emergence of works [50, 57, 59], using other data modalities, such as skeleton, infrared sequence, point cloud, event stream, audio, acceleration, radar. This is mainly due to the development of different kinds of accurate and affordable sensors.

Multi-frame dense optical flow is primarily used to train two-stream CNN [27], of which temporal and spatial stream deals with motion in form of dense optical flow and still video frames respectively. The two-stream 2D CNN framework [44, 47, 49] generally contains two 2D CNN branches taking different input features extracted from the RGB videos for HAR, and the final result is usually obtained through fusion strategies. Additionally, RNNs can be used to analyze temporal data due to the recurrent connections in their hidden layers. Most of the existing methods have adopted gated RNN architectures, such as Long-Short Term Memory (LSTM) [60, 62], to model the long-term temporal dynamics in video sequences.

Plenty of researches [39, 40, 54] have extended 2D CNNs to 3D structures, to simultaneously model the spatial and temporal context information in videos that is crucial for HAR. Transformer [64] is a novel deep learning model leading the machine learning field recently. Transformer is composed of an encoder and a decoder. This design enables Transformer to perform well concerning long-term dependency modeling, multi-modal fusion, and multi-task processing [53, 65].

Skeleton based algorithms encode the trajectories of human body joints, which characterize informative human motions. Therefore, skeleton data is also a suitable modality for HAR. The skeleton data can be acquired by applying pose estimation algorithms on RGB videos [30, 63]. Most of recent works on skeleton based HAR used skeleton data obtained from RGB videos [56].

For human detection, human detectors are mainly divided into one-stage object detectors [21, 28, 29] and two-stage object detectors [9, 10, 31]. One-stage human detectors mainly have two kinds: anchor-based [4, 18, 22, 35] and anchor-free [5, 13, 14, 36, 46]. At present, the more accurate real-time one-stage human detectors are anchor-based EfficientDet [35], YOLOv4 [4], and PP-YOLO [22].

For person counting, early approaches are based on detections [26, 43, 45]. The regression-based approach directly estimates the pedestrian count of the image [17, 23, 24], hindering its extension to localization study. Most recently, person counting via density map estimation has emerged as a promising approach, where the input image is processed to a crowd density map, which is simply integrated to obtain the number of people in a pixel of the image [6, 11, 33, 48].

For sleep detection and fall detection, early approaches are based on SVM [38, 41], KNN [16, 32] and feature-threshold [15, 19], and they use foreground extraction through background subtraction. CNN based approaches [51, 52, 61] obtain features through convolutional network. Skeleton based or human key-points based approaches are also popular [37, 58], and they extract information from the positions, angles and scales of skeletal points and use the information to characterize human actions.

For efficient human action recognition in various scenarios, it is straightforward to do action recognition with the method in Figure 2, where the human detection is applied to an input image followed by the action recognition. Intuitively, different action require different human body information to make a decision. For example, the whole body detection is required for fall detection, the upper body detection for sleep detection, and the part body detection for on-duty detection. Thus, different human detection needs to be setup for different action recognition. In this paper, the general human detection is proposed for supporting various actions and thus used to construct a unified human action recognition framework. Additionally, the method is extended from a single image as input to a video sequence in the computation-effective way.

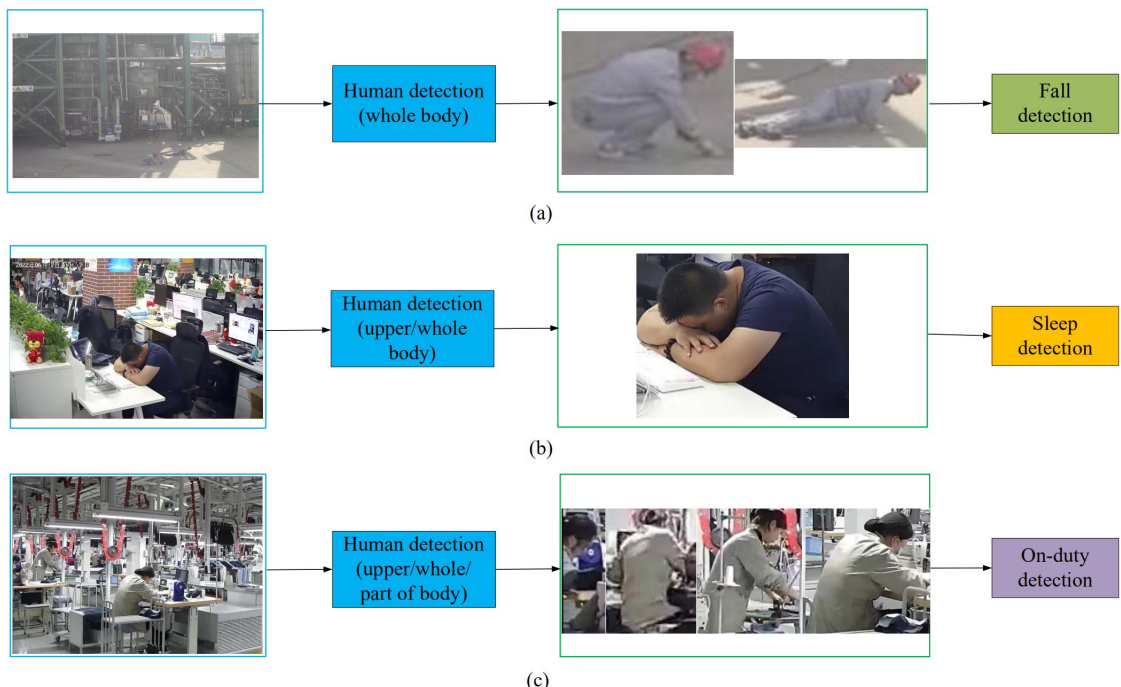

Figure 2. The straightforward human action recognition scheme with different human detection for different action scenario: (a) fall detection, (b) sleep detection, and (c) on-duty detection.

## 3. The Proposed Application-Driven AI Paradigm

### 3.1. System overview

The proposed application-driven AI paradigm for human action recognition, as shown in Figure 3, is composed of two stages: one for multi-form human detection and the other for action classification. In the first stage, the human body regions with three kinds of label are identified from an input image. Then in the second stage, the corresponding kind of human body regions are used to recognize certain action with an action classifier. Thus, for each input image, the system tells the confidence score or action label of each considered action type. For the video sequence composed of temporal image frames, the global action label is acquired by finding the most frequent action label. Instead of dealing with whole video, periodic frames from the video sequence are processed. Also, in most of the cases, a single frame is sufficient for recognition of action in the video. In the present work, we utilize minimum number of frames thus reducing computational time.

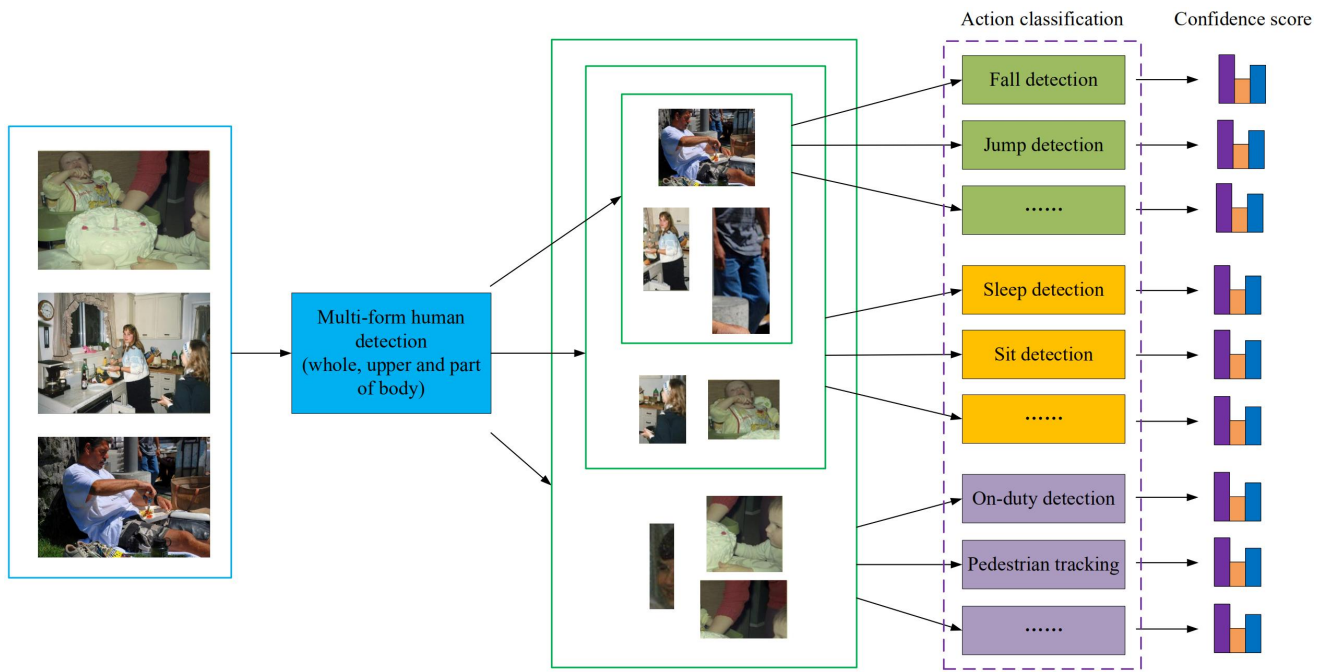

Figure 3. the application-driven AI paradigm for human action recognition. With images as input, the multi-form human detection model produces three kinds of human body regions. Then, various action classifiers are followed corresponding to certain kind of human body regions.

### 3.2. Multi-form human detection

In the proposed multi-form human detection, three kinds of human body regions, i.e., the whole body, upper body and part of body, are localized from an input image. These human body forms may cover nearly all potential application scenarios. For example, such case as on-duty detection or pedestrian tracking requires only the part of body to make a decision, while such case as sleep detection or sit detection often requires the upper body to make a decision. Furthermore, such case as fall detection, jump detection, running detection or standing detection usually requires the whole body to make a decision.

Generally, an object detection model can be used for multi-form human detection. Here, we use YOLOv5 [4], an object detection algorithm that divides an image into a grid system, where each cell in the grid is responsible for detecting objects within itself. For its backbone, we use new CSP-Darknet53 module [12], and use SPPF [7], New CSP-PAN [8] as neck module. Due to the overall trade-off of actual speed requirements, memory consumption and accuracy, we use the m version of the yolov5 model with a resolution of 640x640, which can reach 20fps in a non-GPU deployment environment such as Intel i7-1185g7. And when we use the integrated GPU of Intel CPU, the speed of our algorithm will be increased by 50% to reach 30fps. For training the object detection model, a new dataset is constructed based on COCO2017 [1], which will be presented in detail in Section 4.

### 3.3. Action classification

In action classification, the action classifier decides whether each of the detected human body region corresponds to certain action. Here, various action recognition scenarios can be supported so long as the corresponding action classifiers are provided. Noted that different action scenario may need different kind of human body region. For example, fall detection works on the whole body region, sleep detection on both upper body region and whole body region, while on-duty detection on either part of body region, upper body region or whole body region.

Generally, to design the action classifier, the deep learning based image classification model can be adopted. However, some tricks should be carefully considered when preparing the training dataset. For example, in the case of sleep detection, we need to extend the two classes of sleep/nosleep to three classes of sleep/sit/nosleep or more, because the normal posture of the upper sitting body may be very close to the posture of lying on table sleeping. In another case such as fall detection, the posture of person sitting on bed/chair/sofa may be close to falling, so we need to extend the number of classes of the classifier to reduce false positives as far as possible. In this work, we use resnet18 [66] as the classification model considering the trade-off of accuracy and speed. Differently, for on-duty detection, the person counting based on person detection may be enough to tell whether the number of person is reasonable on work.

## 4. Open-Source of Multi-form Human Detection Dataset

To train the proposed multi-form human detection model, we construct an open-source dataset with two means: Based on COCO2017 [1] dataset, we distinguish three kinds of human body labels, i.e., the whole body, upper body and part of body. In addition to this, we collect and annotate 700 videos from actual applications containing human bodies with different poses in various scenarios to improve the algorithm performance. Totally, we have collected 70k images containing 275,000 human bodies. This dataset covers the whole/upper/partial human bodies with various angles, illuminations, sizes and qualities, and is manually labeled by a number of people in numerous rounds. As shown in Table 1, there are totally 275000 annotations, 102000 for whole bodies, 68000 for upper bodies and 105000 for part of bodies.

TABLE I
HUMAN ANNOTATION STATISTICS

| Annotation strategy | Whole body | Upper body | Part of body |
|---|---|---|---|
| Quantity | 102000 | 68000 | 105000 |

### 4.1. Annotation for whole human bodies

For the whole body's annotation, we set the strategy that the torso part and the legs are mostly not occluded. Specifically, different from the upper body annotation in Section 4.2, the labeling of the whole body focuses more on the torso and lower limbs. Even if the head and hands are occluded, if the torso and lower limbs of the person are visible, we still annotate it as the whole body. Finally, we got 102000 whole body annotations. Some samples are shown in Figure 4, where only the whole bodies are annotated in yellow.

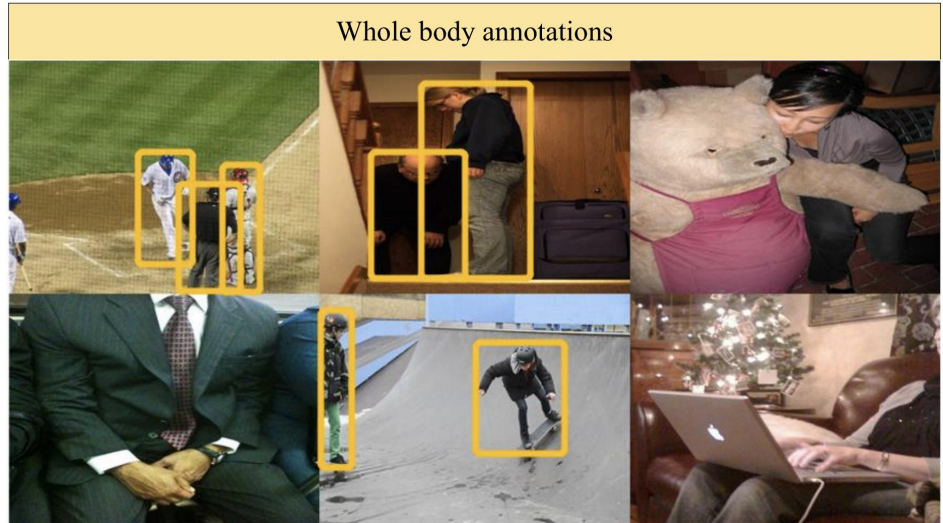
Figure 4. Examples of whole human body annotations. Among these pictures, seven whole human bodies are annotated in yellow.

### 4.2. Annotation for upper human bodies

For the upper body's annotation, we set the strategy that at least the head and shoulder need to be seen. However, according to COCO's labeling standard, a large proportion of the data will be labeled as partial body while not upper body. Therefore, some work needs to be done to update the labels of COCO dataset. Such a training data update finally leaves us with 68000 upper human body labels. Figure 5 shows some examples of upper body annotation in green.

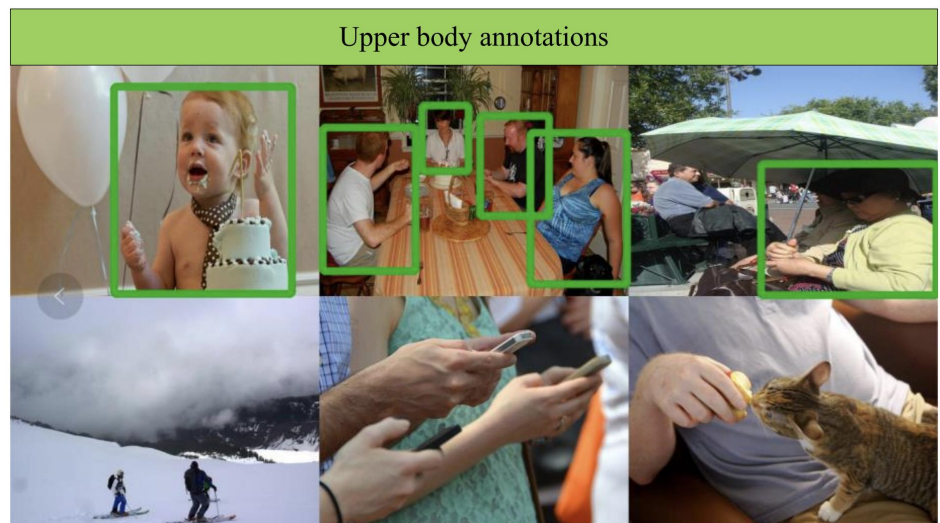
Figure 5. Examples of upper human body annotations. Among these pictures, six upper human bodies are annotated in green.

### 4.3. Annotation for partial human bodies

For COCO2017 dataset, the partial body annotations are obtained after the whole body annotation in Section 4.1 and the upper body annotation in Sections 4.2. For the data collected from the actual applications, we annotate the human bodies according to the same labeling rules. Finally, we obtain a total of 105000 partial human body annotations. Some examples are shown in Figure 6 with partial bodies annotated in blue.

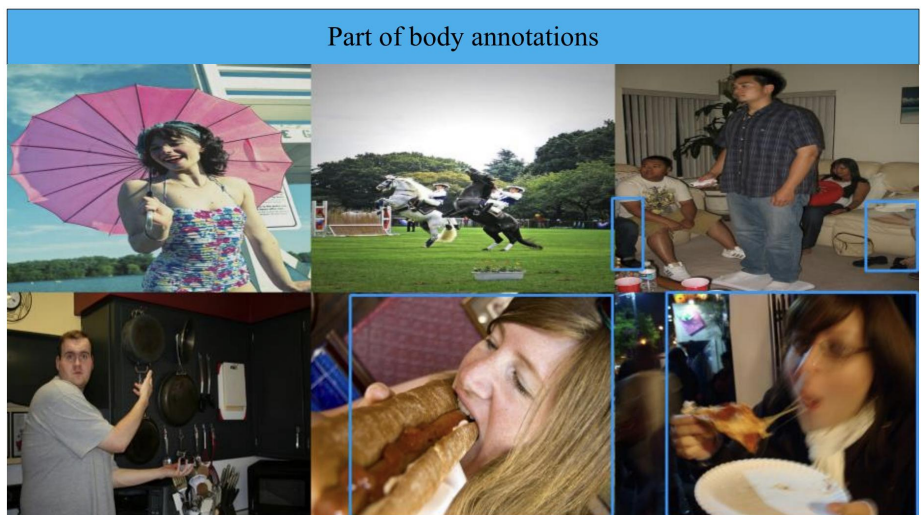
Figure 6. Examples of partial human body annotations. Among these pictures, four partial bodies are annotated in blue.

## 4.4. Annotation for action classification

Based on the annotated human bodies, we setup the action subsets for training action classifiers. As examples, Figure 7 shows the subsets corresponding to standing, jumping, falling, sleeping and sitting, respectively. Based on the subsets, some action classifiers can be trained.

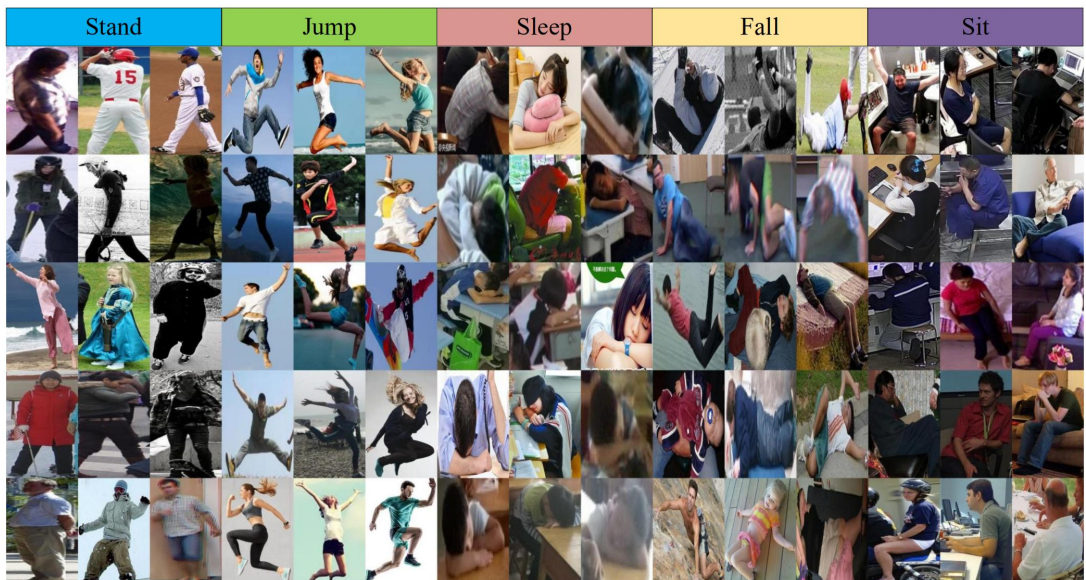
Figure 7. Examples of action subsets: Stand, Jump and Fall denote the standing posture, jumping posture and falling posture respectively based on the whole body, and Sleep and Sit denote the sleeping posture and sitting posture based on the upper body.

## 5. Experiments

### 5.1. Implementation details

For the yolov5 model in multi-form human body detection, we disable mosaic which will produce incomplete human body labels. Copy-paste and random affine (Rotation, Scale, Translation and Shear) are still being used. For rotation parameter, we expand the default (-5, 5) random sampling range to (-90, 90) degrees, which could greatly improve recall rate of inclined human body. For Multi-scale parameter which controls how images scale in training, we set it to (1~1.5x).

For the resnet18 model in action classification, we modify the input 224x224 to 384x128, which is effective for classification model training when the human body is the input. During the training min-batch, we dynamically balance the number of samples in each category.

### 5.2. Experiments on public dataset

**Dataset and Metric:** We evaluate our method on public dataset UR Fall Detection Dataset (UR) [2] and Fall detection Dataset (FDD) [3]. UR Fall Detection Dataset contains 70 (30 falls + 40 activities of daily living) sequences. Out of total datasets of 22636 images, 16794 images can be used for training, 3299 images can be used for validation and 2543 images can be used for the test. When training our network, we follow the original partition. And, we show the quantitative comparison results with sensitivity (true positive rate) and specificity (true negative rate) as metric.

Taking fall detection for example, we compare the original detection-classification method and state of the art methods [51, 62, 93] with our method which using our paradigm. In the original detection-classification method, the original YOLO is adopted for human body detection, which produces the undistinguishable human bodies for action classification. As can be seen from the results in Table II, compared with original detection-classification process, our paradigm greatly improves on the specificity and also improves significantly on the sensitivity. Combining two metrics, our paradigm achieves the best performance compared to other algorithms. With respect to the specificity, it is reasonable that the whole human body is more suitable for fall classification than the upper body or partial body.

TABLE II
EXPERIMENTS ON UR AND FDD DATASETS

| Methods | Sensitivity | Specificity |
|---|---|---|
| Original YOLO + classification | 97.9% | 85.3% |
| SVM [51] | 91.1% | 87.3% |
| CNN [62] | 92.8% | 96.1% |
| CNN + human key points [93] | **99.5%** | 98.7% |
| Our method | 99.4% | **99.2%** |

### 5.3. Experiments on our dataset

We collect video data from actual scenarios for action analysis as our own datasets. The duration of video data is 13 hours, which is made up of 200 video segments. When training our model, two-thirds of the dataset is used for training, while the rest for testing. Figure 8 shows an instance on work status surveillance, which is able to recognize four human actions including standing, sitting, similar to sleeping and sleeping at work.

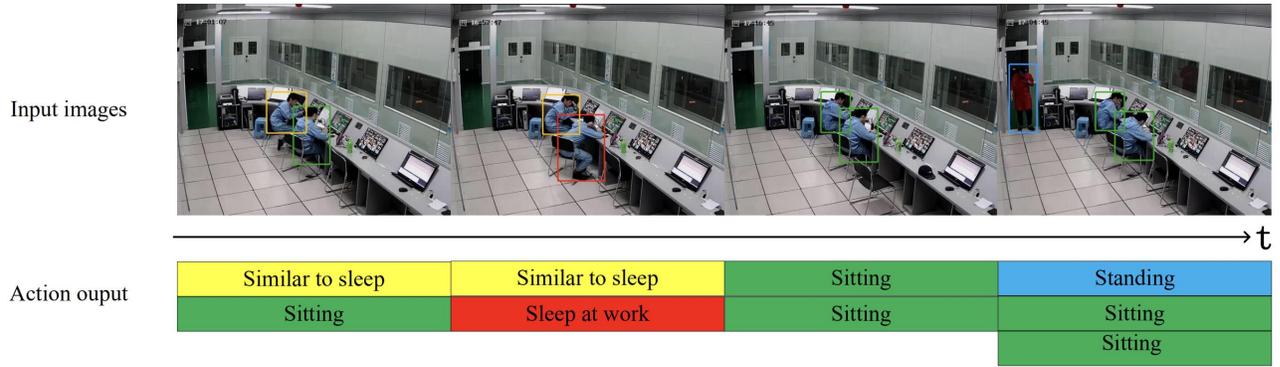

Figure 8. Examples of human action recognition in working environment. In addition to the detection of worker, the system also identifies the action of sleeping at work, and even classifies non-sleeping actions such as standing, sitting, and the one similar to sleeping.

For multi-form human detection, we evaluate our detector with the original YOLO detection. In Table III, we show the quantitative comparison results with precision and recall as metric. Comparing to original detection, we loss little recall rate, but has a boost in precision rate. It shows that to detect human body in three forms respectively can obtains higher precision rate than to detect in only one form.

TABLE III
DETECTORS' EXPERIMENTS ON OUR DATASET

| Methods | Precision | Recall |
| --- | --- | --- |
| Orginal YOLO | 95.9% | **92.2%** |
| Multi-form human detector | **97.1%** | 91.6% |

For action classification, we evaluate our algorithms with the original detection-classification method. In Table IV, we also show the quantitative comparison results of four different action classifications with precision and recall as metric. Comparing to the original detection-classification, we have a slight decrease of recall rate in three applications, and a slight increase in the recall rate of sleep detection. We improve precision rate of all applications, especially for difficult tasks such as fall detection and sleep detection, and the improvement is significant. As is shown that the action classification based on the three forms of human bodies produced by multi-form human body detection obtains higher precision rate than the action classification based on only one form of human body produced by the general person detection such as YOLO.

TABLE IV
CLASSIFIERS' EXPERIMENTS ON OUR DATASET

| Methods | Applications | | | | | | | |
| --- | --- | --- | --- | --- | --- | --- | --- | --- |
| | Fall detection | | Sleep detection | | Jump detection | | On-duty detection | |
| | Precision | Recall | Precision | Recall | Precision | Recall | Precision | Recall |
| Original YOLO + classification | 83.4 | **99.2** | 77.5 | 96.1 | 95.5 | **98.7** | 95.9 | **94.5** |
| Multi-form human detection + classification (our method) | **97.5** | 99.0 | **93.1** | **96.9** | **98.1** | 97.9 | **98.8** | 94.0 |

# 6. Conclusion

We propose an application-driven AI paradigm for human action recognition with a combination of multi-form human detection and action classification. The multi-form human detection model, trained with a new dataset, localizes three forms of human bodies including the whole body, upper body and partial body simultaneously. Based on the localized human bodies, various action classifiers can be used to identify corresponding actions. The new open-source dataset is constructed based on both existing open-source dataset and own-collected data, which is annotated into three forms of human bodies covering nearly all potential applications. Experiment results on public dataset and our dataset confirm that our system has a great value in real-world applications. Additionally, the proposed paradigm has the potentiality on low-cost implementation which will be investigated in future.

## References

[1] T.-Y. Lin, M. Maire, S. Belongie, J. Hays, P. Perona, D.Ramanan, P. Dollar, and C. L. Zitnick, "Microsoft COCO: Common Objects in Context," in European Conferenceon Computer Vision (ECCV), 2014.

[2] Bogdan Kwolek, Michal Kepski, Human fall detection on embedded platform using depth maps and wireless accelerometer, Computer Methods and Programs in Biomedicine, Volume 117, Issue 3, December 2014, Pages 489-501, ISSN 0169-2607

[3] Adhikari, Kripesh, Hamid Bouchachia, and Hammadi Nait Charif. "Activity recognition for indoor fall detection using convolutional neural network." Machine Vision Applications (MVA), 2017 Fifteenth IAPR International Conference on. IEEE, 2017.

[4] Chien-Yao Wang, Alexey Bochkovskiy, and HongYuan Mark Liao. Scaled-YOLOv4: Scaling cross stage partial network. In Proceedings of the IEEE/CVF Conference on Computer Vision and Pattern Recognition (CVPR), pages 13029–13038, 2021.

[5] Kaiwen Duan, Song Bai, Lingxi Xie, Honggang Qi, Qingming Huang, and Qi Tian. CenterNet: Keypoint triplets for object detection. In Proceedings of the IEEE International Conference on Computer Vision (ICCV), pages 6569–6578, 2019. 2.

[6] Bai, H., Wen, S., Gary Chan, S.H.: Crowd counting on images with scale variation and isolated clusters. In: ICCV Workshops, 2019.

[7] Kaiming He, Xiangyu Zhang, Shaoqing Ren, and Jian Sun. Spatial pyramid pooling in deep convolutional networks for visual recognition. IEEE Transactions on Pattern Analysis and Machine Intelligence (TPAMI), 37(9):1904–1916, 2015. 2, 4, 7

[8] Shu Liu, Lu Qi, Haifang Qin, Jianping Shi, and Jiaya Jia. Path aggregation network for instance segmentation. In Proceedings of the IEEE Conference on Computer Vision and Pattern Recognition (CVPR), pages 8759–8768, 2018. 5

[9] Ross Girshick. Fast R-CNN. In Proceedings of the IEEE International Conference on Computer Vision (ICCV), pages 1440–1448, 2015. 2

[10] Ross Girshick, Jeff Donahue, Trevor Darrell, and Jitendra Malik. Rich feature hierarchies for accurate object detection and semantic segmentation. In Proceedings of the IEEE Conference on Computer Vision and Pattern Recognition (CVPR), pages 580–587, 2014. 2

[11] Cao, X., Wang, Z., Zhao, Y., Su, F.: Scale aggregation network for accurate and efficient crowd counting. In: ECCV, 2018.

[12] Chien-Yao Wang, Hong-Yuan Mark Liao, Yueh-Hua Wu, Ping-Yang Chen, Jun-Wei Hsieh, and I-Hau Yeh. CSPNet: A new backbone that can enhance learning capability of cnn. Proceedings of the IEEE Conference on Computer Vision and Pattern Recognition Workshop (CVPR Workshop), 2020. 2, 7

[13] Hei Law and Jia Deng. CornerNet: Detecting objects as paired keypoints. In Proceedings of the European Conference on Computer Vision (ECCV), pages 734–750, 2018. 2

[14] Hei Law, Yun Teng, Olga Russakovsky, and Jia Deng. CornerNet-Lite: Efficient keypoint based object detection. arXiv preprint arXiv:1904.08900, 2019. 2

[15] Yajai, A.; Rodtook, A.; Chinnasarn, K.; Rasmequan, S.; Apichet, Y. Fall detection using directional bounding box. In Proceedings of the 2015 12th International Joint Conference on Computer Science and Software Engineering (JCSSE), Hatyai, Thailand, 22–24 July 2015; pp. 52–57.

[16] Gunale, K.G.; Mukherji, P. Fall detection using k-nearest neighbor classification for patient monitoring. In Proceedings of the 2015 International Conference on Information Processing (ICIP), Pune, India, 16–19 December 2015; pp. 520–524.

[17] Wang, C., Zhang, H., Yang, L., Liu, S., Cao, X.: Deep people counting in extremely dense crowds. In: ACM Multimedia, 2015.

[18] Tsung-Yi Lin, Priya Goyal, Ross Girshick, Kaiming He, and Piotr Dollar. Focal loss for dense object detection. In ´ Proceedings of the IEEE International Conference on Computer Vision (ICCV), pages 2980–2988, 2017. 2, 7

[19] Chong, C.-J.; Tan, W.-H.; Chang, Y.C.; Batcha, M.F.N.; Karuppiah, E. Visual based fall detection with reduced complexity horprasert segmentation using superpixel. In Proceedings of the 2015 IEEE 12th International Conference on Networking, Sensing and Control, Taipei, Taiwan, 9–11 April 2015; pp. 462–467.

[20] W. Lin, M.-T. Sun, R. Poovandran, and Z. Zhang, "Human activity recognition for video surveillance," in ISCAS, 2008.

[21] Wei Liu, Dragomir Anguelov, Dumitru Erhan, Christian Szegedy, Scott Reed, Cheng-Yang Fu, and Alexander C Berg. SSD: Single shot multibox detector. In Proceedings of the European Conference on Computer Vision (ECCV), pages 21–37, 2016. 2

[22] Xiang Long, Kaipeng Deng, Guanzhong Wang, Yang Zhang, Qingqing Dang, Yuan Gao, Hui Shen, Jianguo Ren, Shumin Han, Errui Ding, et al. PP-YOLO: An effective and efficient

implementation of object detector. arXiv preprint arXiv:2007.12099, 2020. 2, 7

[23] Chan, A.B., Vasconcelos, N.: Bayesian poisson regression for crowd counting. In: ICCV, 2009.

[24] Chan, A.B., Vasconcelos, N.: Counting people with low-level features and bayesian regression. TIP, 2012.

[25] M. Lu, Y. Hu, and X. Lu, "Driver action recognition using deformable and dilated faster r-cnn with optimized region proposals," Appl. Intell., vol. 50, no. 4, 2020.

[26] Li, M., Zhang, Z., Huang, K., Tan, T.: Estimating the number of people in crowded scenes by mid based foreground segmentation and head-shoulder detection. In: ICPR, 2008.

[27] J. Shotton, A. Fitzgibbon, M. Cook, T. Sharp, M. Finocchio, R. Moore, A. Kipman, and A. Blake, "Real-time human pose recognition in parts from single depth images," in CVPR, 2011.

[28] Joseph Redmon, Santosh Divvala, Ross Girshick, and Ali Farhadi. You only look once: Unified, real-time object detection. In Proceedings of the IEEE Conference on Computer Vision and Pattern Recognition (CVPR), pages 779–788, 2016. 2

[29] Joseph Redmon and Ali Farhadi. YOLO9000: better, faster, stronger. In Proceedings of the IEEE Conference on Computer Vision and Pattern Recognition (CVPR), pages 7263–7271, 2017. 2

[30] J. Gong, Z. Fan, Q. Ke, H. Rahmani, and J. Liu, "Meta agent teaming active learning for pose estimation," in CVPR, 2022.

[31] Shaoqing Ren, Kaiming He, Ross Girshick, and Jian Sun. Faster R-CNN: Towards real-time object detection with region proposal networks. In Advances in Neural Information Processing Systems (NIPS), pages 91–99, 2015.

[32] De Miguel, K.; Brunete, A.; Hernando, M.; Gambao, E. Home Camera-Based Fall Detection System for the Elderly. Sensors 2017, 17, 2864.

[33] Pham, V.Q., Kozakaya, T., Yamaguchi, O., Okada, R.: Count forest: Co-voting uncertain number of targets using random forest for crowd density estimation. In: ICCV, 2015.

[34] J. Liu, H. Rahmani, N. Akhtar, and A. Mian, "Learning human pose models from synthesized data for robust rgb-d action recognition," IJCV, 2019.

[35] Mingxing Tan, Ruoming Pang, and Quoc V Le. EfficientDet: Scalable and efficient object detection. In Proceedings of the IEEE Conference on Computer Vision and Pattern Recognition (CVPR), 2020. 1, 2, 7

[36] Zhi Tian, Chunhua Shen, Hao Chen, and Tong He. FCOS: Fully convolutional one-stage object detection. In Proceedings of the IEEE International Conference on Computer Vision (ICCV), pages 9627–9636, 2019. 2

[37] Zhang, J.; Wu, C.; Wang, Y. Human Fall Detection Based on Body Posture Spatio-Temporal Evolution. Sensors 2020, 20, 946.

[38] Juang, L.H.; Wu, M.N. Fall Down Detection Under Smart Home System. J. Med. Syst. 2015, 39, 107–113.

[39] S. Ji, W. Xu, M. Yang, and K. Yu, "3d convolutional neural networks for human action recognition," TPAMI, vol. 35, no. 1, 2012.

[40] H. Zhang, L. Zhang, X. Qui, H. Li, P. H. Torr, and P. Koniusz, "Few-shot action recognition with permutation-invariant attention," in ECCV, 2020.

[41] Aslan, M.; Sengur, A.; Xiao, Y.; Wang, H.; Ince, M.C.; Ma, X. Shape feature encoding via Fisher Vector for efficient fall detection in depth-videos. Appl. Soft Comput. 2015, 37, 1023–1028.

[42] R. Poppe, "A survey on vision-based human action recognition," Image Vis Comput, 2010.

[43] Lin, Z., Davis, L.S.: Shape-based human detection and segmentation via hierarchical part-template matching. TPAMI, 2010.

[44] K. Simonyan and A. Zisserman, "Two-stream convolutional networks for action recognition in videos," in NeurIPS, 2014.

[45] Rabaud, V., Belongie, S.: Counting crowded moving objects. In: CVPR, 2006.

[46] Xingyi Zhou, Dequan Wang, and Philipp Krahenb¨uhl. Ob-¨ jects as points. In arXiv preprint arXiv:1904.07850, 2019. 2

[47] A. Karpathy, G. Toderici, S. Shetty, T. Leung, R. Sukthankar, and L. Fei-Fei, "Large-scale video classification with convolutional neural networks," in CVPR, 2014.

[48] Zhang, Y., Zhou, D., Chen, S., Gao, S., Ma, Y.: Single-image crowd counting via multi-column convolutional neural network. In: CVPR, 2016.

[49] H. Bilen, B. Fernando, E. Gavves, and A. Vedaldi, "Action recognition with dynamic image networks," TPAMI, vol. 40, no. 12, 2017.

[50] J. Donahue, L. Anne Hendricks, S. Guadarrama, M. Rohrbach, S. Venugopalan, K. Saenko, and T. Darrell, "Long-term recurrent convolutional networks for visual recognition and description," in CVPR, 2015.

[51] Li, X.; Pang, T.; Liu, W.; Wang, T. Fall detection for elderly person care using convolutional neural networks. In Proceedings of the 2017 10th International Congress on Image and Signal Processing, BioMedical Engineering and Informatics (CISP-BMEI), Shanghai, China, 14–16 October 2017; pp. 1–6.

[52] Fan, Y.; Levine, M.D.; Wen, G.; Qiu, S. A deep neural network for real-time detection of falling humans in naturally occurring scenes. Neurocomputing 2017, 260, 43–58.

[53] S. Khan, M. Naseer, M. Hayat, S. W. Zamir, F. S. Khan, and M. Shah, "Transformers in vision: A survey," ACM Comput. Surv., 2021.

[54] D. Tran, L. Bourdev, R. Fergus, L. Torresani, and M. Paluri, "Learning spatiotemporal features with 3d convolutional networks," in ICCV, 2015.

[55] J. Liu, A. Shahroudy, M. L. Perez, G. Wang, L.-Y. Duan, and A. C. Kot, "Ntu rgb+d 120: A large-scale benchmark for 3d human activity understanding," TPAMI, 2020.

[56] S. Yan, Y. Xiong, and D. Lin, "Spatial temporal graph convolutional networks for skeleton-based action recognition," in AAAI, 2018.

[57] J. Liu, A. Shahroudy, D. Xu, A. C. Kot, and G. Wang, "Skeleton-based action recognition using spatio-temporal lstm network with trust gates," TPAMI, vol. 40, no. 12, 2018.

[58] Xu, Q.; Huang, G.; Yu, M.; Guo, Y.; Huang, G. Fall prediction based on key points of human bones. Phys. A Stat. Mech. its Appl. 2020, 540, 123205.

[59] H. Rahmani and A. Mian, "3d action recognition from novel viewpoints," in CVPR, 2016.

[60] W. Du, Y. Wang, and Y. Qiao, "Rpan: An end-to-end recurrent pose-attention network for action recognition in videos," in ICCV, 2017.

[61] Rahnemoonfar, M.; Alkittawi, H. Spatio-temporal convolutional neural network for elderly fall detection in depth video cameras. In Proceedings of the 2018 IEEE International Conference on Big Data (Big Data), Seattle, WA, USA, 10–13 December 2018; pp. 2868–2873.

[62] L. Sun, K. Jia, K. Chen, D.-Y. Yeung, B. E. Shi, and S. Savarese, "Lattice long short-term memory for human action recognition," in ICCV, 2017.

[63] K. Sun, B. Xiao, D. Liu, and J. Wang, "Deep high-resolution representation learning for human pose estimation," in CVPR, 2019.

[64] A. Vaswani, N. Shazeer, N. Parmar, J. Uszkoreit, L. Jones, A. N. Gomez, Ł. Kaiser, and I. Polosukhin, "Attention is all you need," NeurIPS, 2017.

[65] K. Han, Y. Wang, H. Chen, X. Chen, J. Guo, Z. Liu, Y. Tang, A. Xiao, C. Xu, Y. Xu, et al., "A survey on vision transformer," TPAMI, 2022.

[66] He K, Zhang X, Ren S, et al. "Deep residual learning for image recognition," in CVPR, 2016.